\documentclass{article}

 \usepackage[preprint]{neurips_2026}

\usepackage[utf8]{inputenc} 
\usepackage[T1]{fontenc}    
\usepackage{hyperref}       
\usepackage{url}            
\usepackage{booktabs}       
\usepackage{amsfonts}       
\usepackage{nicefrac}       
\usepackage{microtype}      
\usepackage{xcolor}         

\usepackage{wrapfig}
\usepackage{minted}

\usepackage{amsmath}
\usepackage{amssymb}
\usepackage{mathtools}
\usepackage{amsthm}
\usepackage{subcaption}
\usepackage{algorithm}
\usepackage{algpseudocode}
\usepackage{enumitem}

\usepackage[textsize=footnotesize]{todonotes}
\usepackage{colortbl}
\definecolor{lightgray}{gray}{0.92}
\definecolor{midgray}{gray}{0.88}
\definecolor{darkgray}{gray}{0.7}
\usepackage{multirow}
\usepackage{graphicx}
\usepackage{xcolor}
\usepackage[table]{xcolor}

\theoremstyle{plain}

\theoremstyle{definition}

\theoremstyle{remark}

\usepackage{amsmath,amsfonts,bm}
\usepackage{xcolor}

\definecolor{deepyellow}{RGB}{200,160,50}
\definecolor{deepgreen}{RGB}{40,135,100}
\definecolor{deepred}{RGB}{190,80,80}

\def\eqref#1{Eq.~(\ref{#1})}

\def\1{\bm{1}}

\DeclareMathAlphabet{\mathsfit}{\encodingdefault}{\sfdefault}{m}{sl}
\SetMathAlphabet{\mathsfit}{bold}{\encodingdefault}{\sfdefault}{bx}{n}

\def\gF{{\mathcal{F}}}

\def\gP{{\mathcal{P}}}

\def\gT{{\mathcal{T}}}

\title{Towards One-for-All Robustness \\Across a Continuum of  Threat Levels}

\author{%
  Zhichao Hou \qquad Xiaorui Liu \thanks{Corresponding author.}\\
North Carolina State University\\
  \texttt{zhou4@ncsu.edu, xliu96@ncsu.edu}
}

\begin{document}

\maketitle


\begin{abstract}


Adversarially robust models often overfit to a specific attack budget, necessitating multiple specialized models for diverse and dynamic adversarial environments, a strategy that becomes fundamentally intractable as the threat space grows. This raises an open challenge: can we achieve strong robustness across a continuum of threat levels within a single model? 
We propose the Threat Conditional Network (TCN), grounded in a representation factorization framework that decomposes representation learning into a threat-invariant shared backbone and a lightweight threat-conditional adaptor. TCN conditions a single model on the perturbation level via Fourier-based embeddings and channel-wise affine modulation, and is trained against a distribution over perturbation budgets, enabling flexible and seamless adaptation across an infinite continuum of threat levels during inference. Extensive experiments on CIFAR-10, CIFAR-100, and Tiny-ImageNet show that TCN matches or surpasses a full ensemble of budget-specialized models with a single set of parameters, generalizes to unseen perturbation budgets, and transfers robustly under mismatched threat conditions, with only 4.6\% parameter overhead. These contributions chart a promising path toward adaptive and generalizable robustness in dynamic and diverse threat environments.


\end{abstract}



\vspace{-0.1in}
\section{Introduction}
\label{sec:intro}


Modern machine learning systems are increasingly deployed in diverse and 
dynamic environments, yet prevailing training paradigms assume 
a static data distribution, which  rarely holds in practice. 
This fundamental mismatch between training and deployment conditions leads 
to substantial performance degradation when the underlying distribution 
shifts at test time~\citep{koh2021wilds, li2023oodrobustbench}.
This phenomenon is particularly pronounced in adversarial robustness. In practice, different deployment environments correspond to different threat types or levels.  Models trained under a specific threat exhibit strong specialization: each model performs well in its training seen threat but degrades significantly under mismatched conditions~\citep{kang2019testing, tramer2019adversarial}.


To address this, numerous endeavors have been proposed, including 
redesigning the training objective to achieve a better accuracy-robustness 
trade-off~\citep{zhang2019theoretically, wang2019improving}, scaling model 
capacity~\citep{rice2020overfitting}, exposing models to multiple threat 
types during training~\citep{tramer2019adversarial, maini2020adversarial, 
laidlaw2020perceptual}, and ensemble-based 
methods~\citep{pang2019improving, strauss2017ensemble, cheng2021mixture}. 
However, performance gains across diverse threats remain limited 
across all these approaches, as the inherent conflict among different 
threat objectives fundamentally constrains generalization.

Therefore,  the conventional method to train and maintain a collection of specialized models 
becomes necessary when real-world systems demand optimal performance 
across diverse threat conditions. While managing a small number of models 
may be feasible, this approach becomes fundamentally intractable as the 
threat space grows increasingly large or even extends to a continuous 
spectrum, as illustrated in Section~\ref{sec:pre}. This raises a 
fundamental open challenge: \emph{Can we achieve strong and consistent 
robustness across a continuum of threat levels within a single model?}


To this end, we propose a novel representation 
factorization framework that explicitly decomposes the embedding mapping
into a threat-invariant shared component and a lightweight 
threat-conditional component. Building on this principle, we design the 
Threat Conditional Network (TCN), a single unified model that 
achieves smooth and consistent adaptation across a continuous spectrum of 
threat levels without the overhead of maintaining multiple specialized 
models. Our contributions are summarized as follows. 
First, we conduct a systematic preliminary study on adversarial defenses across various threat settings, revealing the inherent inconsistency and scalability limitations of existing approaches (Section~\ref{sec:pre}). 
Second, starting from a representation learning perspective, we reveal the underlying limitations of existing methods and propose a novel representation factorization framework. Building on this, we introduce a unified threat-conditional network (TCN) that  enables a single model to adapt and generalize robustly across a continuum of perturbation budgets (Section~\ref{sec:method}). 
Finally, we present extensive experiments to demonstrate that TCN achieves strong one-for-all robustness and across-threat transferability. Additional ablation studies further validate the 
working mechanism and practical advantages of our design (Section~\ref{sec:exp}). Together, these contributions chart a promising path toward adaptive and generalizable robustness in dynamic and diverse threat environments.

\vspace{-0.1in}
\section{Challenges in Scaling Adversarial Defenses Across Threat Levels}
\vspace{-0.1in}

\label{sec:pre}







Most adversarial defenses are tightly coupled to a specific, predefined threat, leading to significant performance degradation when evaluated against 
out-of-scope attacks
Numerous efforts have been devoted to improving generalization across threat levels, yet each line of work faces notable limitations: (1) 
One direction focuses on redesigning the training objective. Methods such as 
TRADES~\citep{zhang2019theoretically} and MART~\citep{wang2019improving} 
introduce new training loss formulations to balance accuracy and robustness, 
yet yield marginal improvement in generalization across threat budgets.  As 
shown in~\cite{gowal2020uncovering}, vanilla adversarial training (AT) can 
achieve comparable trade-offs simply by adjusting the training perturbation 
radius.
(2) Scaling model capacity is another promising direction. Larger models have been 
shown to improve robustness across budgets~\citep{rice2020overfitting}, but 
performance tends to saturate as model width increases, and the growing 
computational cost makes this approach impractical for resource-constrained 
settings.
(3) Another direction exposes the model to multiple threat types during 
training~\citep{tramer2019adversarial, maini2020adversarial, 
cai2018curriculum}, enabling it to handle heterogeneous attacks 
simultaneously. However, the inherent tension among diverse threat 
objectives inevitably leads to performance trade-offs across attack types.
(4) Ensemble-based methods~\citep{pang2019improving, strauss2017ensemble, 
cheng2021mixture} have also been explored to improve cross-budget 
generalization. However, the performance gains remain limited, and these 
methods require training a collection of models simultaneously, which does 
not scale well in practice.

\begin{wrapfigure}{r}{0.6\textwidth}
    \centering
    \vspace{-0.1in}
    \includegraphics[width=0.59\textwidth]{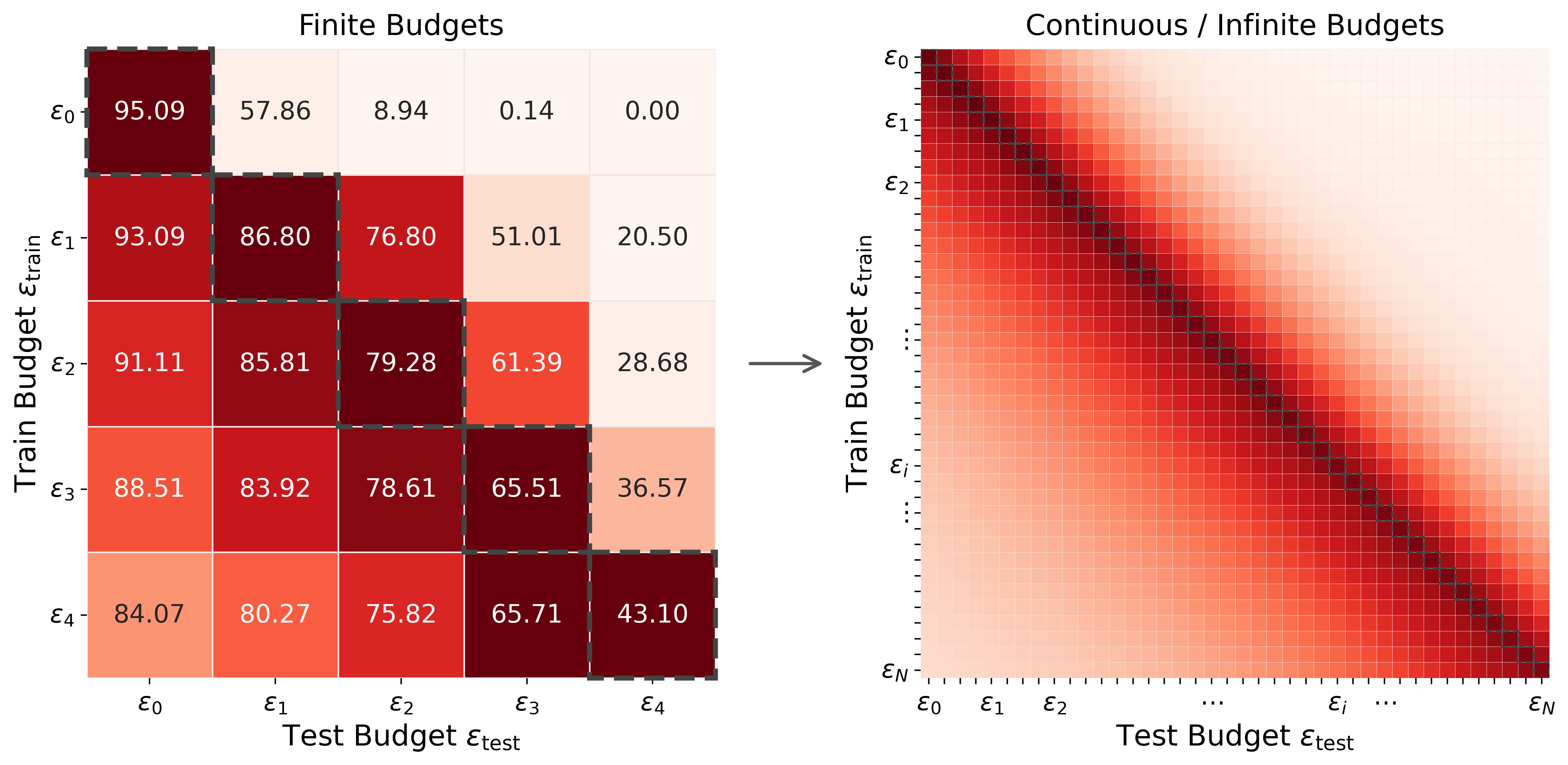}
    \caption{Diagonal dominance pattern under a finite (left) and continuous 
    (right) threat space.}
    \label{fig:finite_to_infinite}
    \vspace{-0.1in}
\end{wrapfigure}

In real-world scenarios, however, models are often deployed under diverse and unpredictable 
threat levels, ideally requiring a dedicated model optimized for each specific 
regime. Due to the inherent performance conflict across budgets in all existing 
defenses, the conventional solution remains training and maintaining a separate 
collection of models. To better illustrate the necessity and limitations of this conventional approach, we conduct the following preliminary study.
We train PGD-AT models under each budget in $\{\epsilon_0, \epsilon_1, 
\epsilon_2, \epsilon_3, \epsilon_4\} = \{\frac{0}{255}, \frac{1}{255}, 
\frac{2}{255}, \frac{4}{255}, \frac{8}{255}\}$, where each model is trained 
at a single budget and evaluated across all budgets. As shown in 
Figure~\ref{fig:finite_to_infinite} (Left), a clear \emph{diagonal dominance} 
pattern emerges: within each test budget (column), the highest accuracy is 
nearly always achieved by the model trained at the matching perturbation level. 
This confirms that: (1) robustness learned at one budget generalizes poorly to 
others; (2) stronger adversarial training sacrifices clean accuracy and low-budget robustness; and (3) weaker training fails to defend against stronger attacks.

Consequently, while the conventional remedy of training one model per threat 
level is effective in isolation, it introduces fundamental scalability 
limitations: training cost scales 
linearly with the number of threat levels, quickly becoming prohibitive in 
large-scale systems. Although maintaining a finite set of models may be manageable for a 
small number of discrete budgets, the problem becomes fundamentally intractable 
as the threat space grows increasingly fine-grained or extends to a continuous 
spectrum (Figure~\ref{fig:finite_to_infinite} Right). This motivates the need for a single model capable of adapting and generalizing robustly across all threat levels.

\section{Threat Conditional Representation Learning}
\label{sec:method}

In this section, we start from a representation learning perspective to analyze the inherent conflict in conventional training method. Building upon this, we propose a novel threat conditional network, along with a distributional adversarial training strategy, to address the open challenge above.

\subsection{Representation Factorization}

\begin{figure}[h!]
    \centering
    \vspace{-0.15in}
    \includegraphics[width=0.99\linewidth]{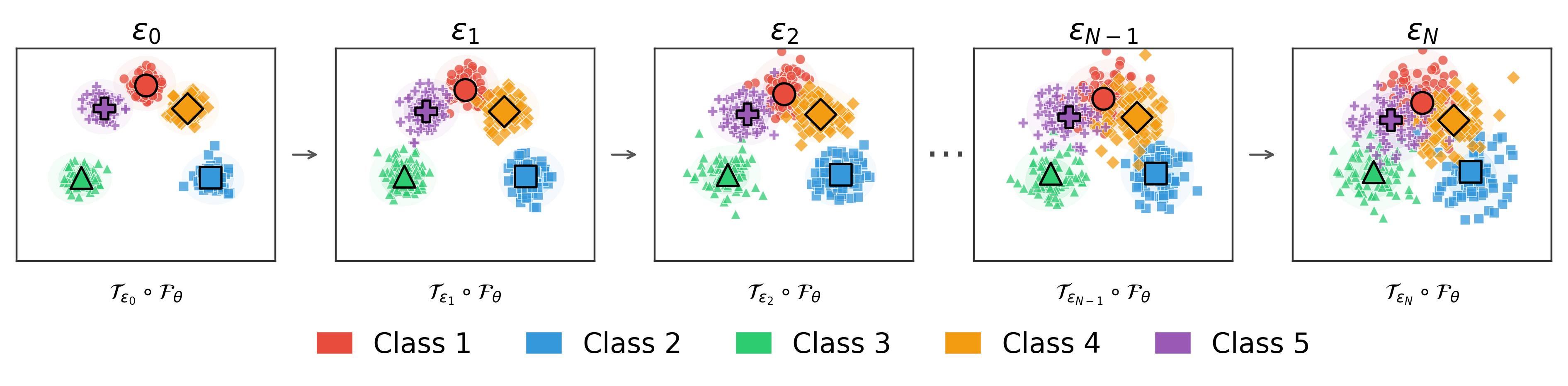}
    \vspace{-0.1in}
\caption{Threat-conditional embedding structure across threat levels. The topological arrangement of embedding remains consistent across all levels.}
    \label{fig:embedding_change}
    \vspace{-0.13in}
\end{figure}

As demonstrated in Section~\ref{sec:pre}, defending against a continuous spectrum of threat levels with a single model is highly non-trivial.
From a representation learning perspective, features learned under different threat levels are inherently conflicting: stronger adversarial training enforces smoother, more invariant representations at the cost of fine-grained discriminability, while weaker training preserves discriminability but lacks robustness. Although existing methods such as multi-threat training or ensemble methods offer partial relief, the underlying representation conflict persists, forcing the learned features to settle at a compromised intermediate level, which inevitably trades off performance across threat levels.

The conventional solution is to maintain a collection of specialized models, each tailored to a specific threat level. However, this quickly becomes intractable as the threat space grows, requiring either a discrete set of parameters $\{\theta_i\}_{i=1}^N$, or even an infinite family $\{\theta(\epsilon)\}_{\epsilon\in\mathbb{R}^+}$, imposing substantial memory and computational overhead. We hypothesize that such large model collections entail significant redundancy: as illustrated in Figure~\ref{fig:embedding_change}, representations across threat levels share a common topological structure, differing only through level-specific adjustments.
This observation motivates a more principled factorization. Let $\gF_\theta : \mathcal{X} \to \mathbb{R}^d$ denote the backbone embedding mapping with parameter $\theta$. Rather than learning a separate $\gF_{\theta(\epsilon)}$ for each $\epsilon$, we propose to decompose the threat-level dependency as: $\gF_{\theta(\epsilon)} \approx \mathcal{T}_\epsilon \circ \gF_\theta,$
where $\gF_\theta$ is a shared backbone and $\mathcal{T}_\epsilon: \mathbb{R}^d \to \mathbb{R}^d$ is a lightweight threat-conditional operator that adapts the shared representation to threat level $\epsilon$. Under this factorization, the full family $\{\gF_{\theta(\epsilon)}\}_{\epsilon \in \mathbb{R}^+}$ is replaced by a single backbone $\mathcal{F}_\theta$ paired with a 
threat-conditional adaptor $\mathcal{T}_\epsilon$ (a lightweight network 
that takes $\epsilon$ as input), reducing the parameter overhead from $O(N|\mathcal{F}|)$ to 
$O(|\mathcal{F}| + |\mathcal{T}|)$, where $|\cdot|$ denotes the number 
of parameters and $|\mathcal{T}| \ll |\mathcal{F}|$.

\subsection{TCN: Threat Conditional Network}

\begin{wrapfigure}{r}{0.38\textwidth}
    \centering
    \vspace{-0.06in}
\includegraphics[width=0.32\textwidth]{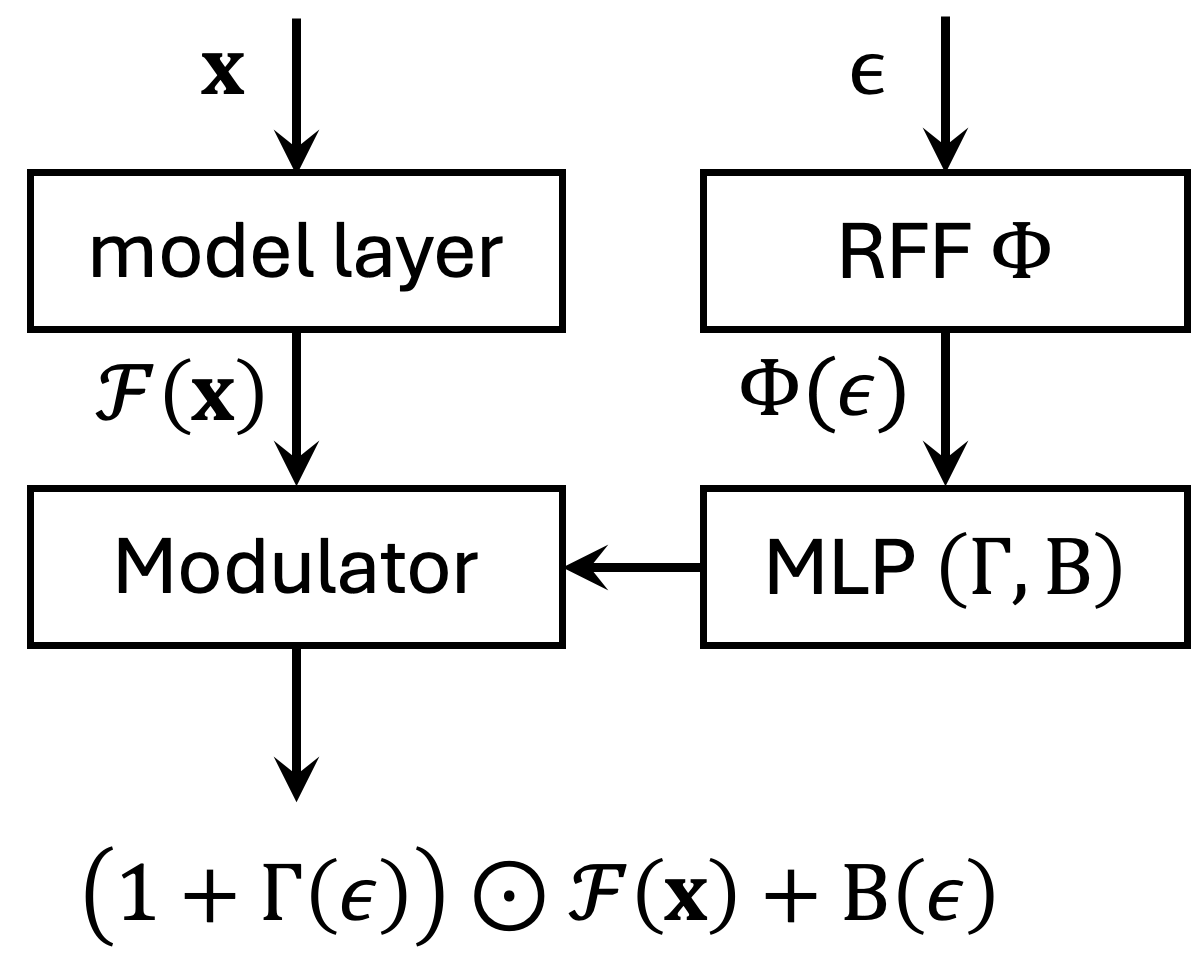}
    \caption{Threat conditional learning.}
    \label{fig:noise_condition_learning}
\end{wrapfigure}

With the decomposition $\gF_{\theta(\epsilon)} \approx \mathcal{T}_\epsilon \circ \gF_\theta$, any backbone architecture can serve as $\gF_\theta$, reducing the problem to efficiently parameterizing the operator family $\{\mathcal{T}_\epsilon\}_{\epsilon \in \mathbb{R}^+}$.
To this end, we propose a threat-conditional learning module, illustrated in Figure~\ref{fig:noise_condition_learning}. Given input $\mathbf{x}$ and threat level $\epsilon$, the module intercepts the intermediate backbone representation and applies a lightweight conditional transformation to adapt it to the specified threat level. The architectural details are described below.

\textbf{Threat-level Embedding.}
To enable expressive conditioning on the scalar $\epsilon$, we adopt a Fourier feature embedding~\citep{rahimi2007random}:
$$
\Phi(\epsilon) = [\sin(2\pi \boldsymbol{\omega} \epsilon), \cos(2\pi \boldsymbol{\omega} \epsilon)] \in \mathbb{R}^d,\; \boldsymbol{\omega} \sim \mathcal{N}(\boldsymbol{0}, \sigma^2 I_{d/2}).$$
This embedding maps $\epsilon$ into a high-dimensional representation space, allowing the model to approximate complex, non-linear dependencies on the perturbation level. 
From a kernel perspective, this can be viewed as enabling a rich class of stationary functions over $\epsilon$, facilitating smooth interpolation across various perturbation strengths.

\begin{figure}[h!]
\centering
\includegraphics[width=0.99\linewidth]{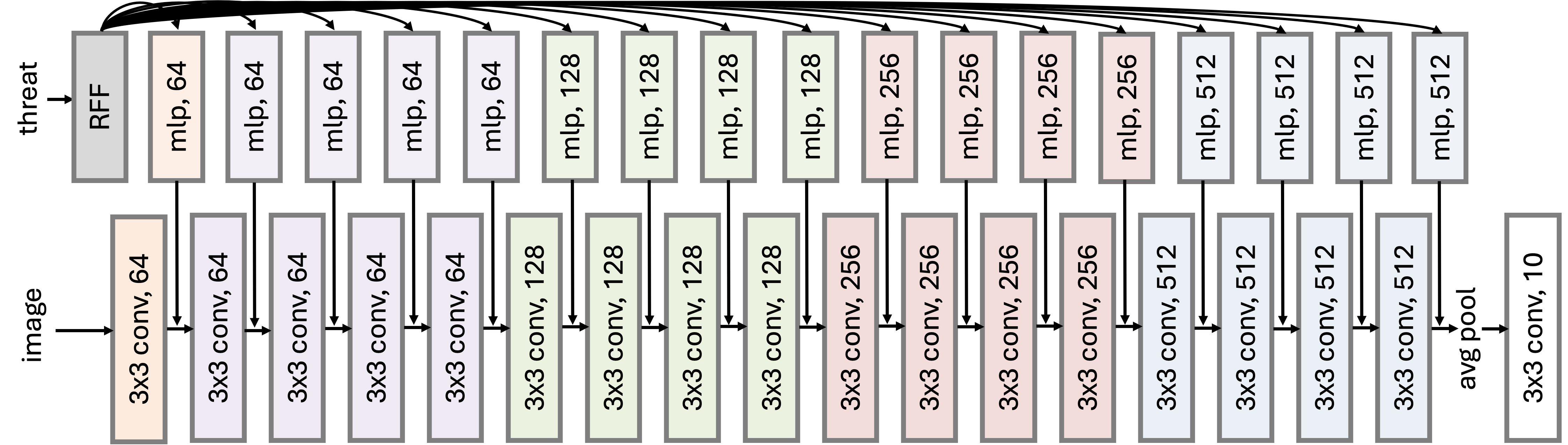}
\caption{Threat-Conditional Network architecture with an 18-layer plain convolutional backbone. }
\vspace{-0.1in}
\label{fig:tcn_arch}
\end{figure}

\textbf{Channel-wise Conditional Modulation.}
Given the embedding $\Phi(\epsilon)$, we generate conditioning parameters through a lightweight  network:
$(\Gamma(\epsilon), \mathrm{B}(\epsilon)) = \mathbf{MLP}(\Phi(\epsilon)).$
These parameters are then injected into intermediate feature activations through a channel-wise affine modulation as inspired by~\citet{perez2018film}. So the threat-conditional operator can be parameterized as 
\begin{equation}
\mathcal{T}_\epsilon\circ \gF: \mathbb{R}^{d_{\mathrm{in}}} \times \mathbb{R} \to \mathbb{R}^{d_{\mathrm{out}}}, \quad (\mathbf{x}, \epsilon) \mapsto (1+\Gamma(\epsilon)) \odot \mathcal{F}(\mathbf{x}) + \mathrm{B}(\epsilon),
\end{equation}
where $\gF(\mathbf{x})$ denotes an intermediate feature tensor and $\odot$ denotes element-wise multiplication. 
This mechanism allows the same backbone network to realize different feature geometries under different threat levels, without introducing separate models for each threat budget. 
Compared with training multiple independent robust classifiers, such conditional modulation offers a more parameter-efficient way to encode perturbation-specific behavior while preserving shared semantic structure across regimes. The threat-conditional operator $\mathcal{T}_\epsilon$ can be inserted after any intermediate layer of a given backbone. Figure~\ref{fig:tcn_arch} provides an overview of the resulting Threat-Conditional Network (TCN) built upon an 18-layer plain convolutional network.

\subsection{Distributional Adversarial Training across Threat Levels}

Tailored to  the threat-conditional network, the training objective should also move beyond a single fixed perturbation radius. In realistic deployment scenarios, the attack strength faced by the model is rarely known in advance and may vary across inputs or environments. It is therefore more principled to optimize the model with respect to a distribution over threat levels. Formally, let $\mathcal{P}$ be a distribution over perturbation radii. We train TCN by solving the following objective:
\begin{equation}
\min_{\theta}
\mathbb{E}_{(\mathbf{x},y)\sim \mathcal{D}}
\left[
\mathbb{E}_{\epsilon \sim \mathcal{P}}
\left[
\max_{\mathbf{x}' \in \mathcal{B}(\mathbf{x}, \epsilon)}
\mathcal{L}\bigl(\mathcal{T}_\epsilon\circ\mathcal{F}_\theta(\mathbf{x}'), y\bigr)
\right]
\right],
\end{equation}
where $\gP$ can be any threat level distribution,  $\mathcal{B}(\mathbf{x}, \epsilon)$ denotes the admissible perturbation set under budget $\epsilon$, and $\mathcal{L}$ is the classification loss.
This objective can be viewed as adversarial training over a continuum of perturbation budgets rather than at a single fixed radius. By sampling $\epsilon$ from $\mathcal{P}$ during training, TCN is encouraged to learn a adaptive threat-conditional decision rule  that remains effective under varying attack strengths, instead of overfitting to one particular threat level.

\section{Experiments}
\label{sec:exp}

In this section, we conduct a comprehensive evaluation of our method, including one-for-all robustness, overall performance comparison, transferability analysis, and ablation studies.

\subsection{Experimental Settings}
\label{sec:exp_set}

\textbf{Datasets and Models.}
We conduct experiments on three widely used benchmarks: CIFAR-10~\citep{krizhevsky2009learning},
CIFAR-100~\citep{krizhevsky2009learning}, and Tiny-ImageNet~\citep{le2015tiny}.
We adopt ResNet-18 and WideResNet-28-10~\citep{he2016deep} as backbone architectures.
ResNet-18 is used as the default unless otherwise specified.

\textbf{Baselines and Evaluation.} 
We compare against several groups of baselines: (1) single-model AT baselines, 
including vanilla PGD-AT~\citep{madry2017towards} trained with both fixed and 
uniform perturbation budgets, as well as representative adversarial training 
methods TRADES~\citep{zhang2019theoretically}, MART~\citep{wang2019improving}, 
and NuAT~\citep{sriramanan2021towards}; and (2) multi-model baselines, 
including MoRE~\citep{cheng2021mixture} and ensemble models.
We evaluate robustness under a diverse set of adversarial attacks, including
FGSM~\citep{goodfellow2014explaining}, PGD~\citep{madry2017towards},
and AutoAttack~\citep{croce2020reliable}. We use PGD-20 as the default attack unless otherwise specified.


\textbf{Training Details.}
All models are trained for 200 epochs with a batch size of 128, momentum of 0.9,
and weight decay of $2\times10^{-5}$. We use SGD with an initial learning rate of 0.1,
decayed by a factor of 10 at epochs 100 and 150.
In our experiments, we model the training threat distribution $\mathcal{P}$ as a \emph{clean-biased uniform distribution}: a mixture that places elevated probability mass on the clean budget $\epsilon_0$ while distributing the remainder uniformly across the remaining threat levels:
$\mathcal{P} = p \cdot \delta_{\epsilon_0} + (1-p)\,\mathrm{Unif}\{\epsilon_1,\ldots,\epsilon_N\},
$
where $p = \frac{M}{M+N}$. Here, $N+1$ is the total number of threat levels under consideration, and $M$ is a hyperparameter that governs the relative sampling weight assigned to clean examples; larger $M$ increasingly biases training toward the clean regime.
We provide the ablation study on $p$ in Appendix~\ref{sec:clean_bias_ablation}.

\subsection{One-for-all Defense}

\begin{table}[h!]
    \centering
    \vspace{-0.2in}
    \caption{Accuracy (\%) across different adversarial training and test budgets.
The best performance at each test budget is predominantly achieved by the model trained
under the same budget, revealing a clear diagonal dominance pattern.
TCN approximates this diagonal performance with a single model. }
    \vspace{0.1in}
    \setlength{\tabcolsep}{3pt} 
    \renewcommand{\arraystretch}{1.3}
\resizebox{1.0\textwidth}{!}{%
    \begin{tabular}{c|c|ccccc|cccccc}
    \toprule
    &\textbf{Backbone}&\multicolumn{5}{c|}{\textbf{ResNet18}}&\multicolumn{5}{c}{\textbf{WideResNet-28-10}}\\
    \midrule
\textbf{Dataset}&\textbf{Method} $\backslash$ \textbf{Budget}&  $\mathbf{\frac{0}{255}}$&$\mathbf{\frac{1}{255}}$&$\mathbf{\frac{2}{255}}$&$\mathbf{\frac{4}{255}}$&$\mathbf{\frac{8}{255}}$&$\mathbf{\frac{0}{255}}$&$\mathbf{\frac{1}{255}}$&$\mathbf{\frac{2}{255}}$&$\mathbf{\frac{4}{255}}$&$\mathbf{\frac{8}{255}}$
\\\midrule
  &AT w/ $\epsilon=\frac{0}{255}$&\cellcolor{gray!20} 95.09 & 57.86 & 8.94 & 0.14 & 0.00&\cellcolor{gray!20}95.37 & 50.41 & 5.44 & 0.00 & 0.00\\
  &AT w/ $\epsilon=\frac{1}{255}$&93.09 & \cellcolor{gray!20}86.80 & 76.80 & 51.01 & 20.50&94.67 & \cellcolor{gray!20}88.55 & 78.66 & 50.20 & 14.12\\
 \textbf{CIFAR10} &AT w/ $\epsilon=\frac{2}{255}$&91.11 & 85.81 & \cellcolor{gray!20} 79.28 & 61.39 & 28.68&93.22 & 88.50 & \cellcolor{gray!20}81.91 & 63.93 & 28.29\\
  &AT w/ $\epsilon=\frac{4}{255}$&88.51 & 83.92 & 78.61 & \cellcolor{gray!20} 65.51 & 36.57&90.91 & 86.72 & 81.90 & \cellcolor{gray!20} 68.66 & 39.79\\
  &AT w/ $\epsilon=\frac{8}{255}$&84.07 & 80.27 & 75.82 & 65.71 & \cellcolor{gray!20} 43.10&86.40 & 82.57 & 78.13 & 68.15 & \cellcolor{gray!20} 45.09\\
  &TCN (Ours)&\cellcolor{gray!30}93.85 & \cellcolor{gray!30} 87.46 & \cellcolor{gray!30} 80.49 & \cellcolor{gray!30} 66.36 & \cellcolor{gray!30} 45.16& \cellcolor{gray!30} 94.56 & \cellcolor{gray!30} 88.49 & \cellcolor{gray!30} 82.32 & \cellcolor{gray!30} 68.87 & \cellcolor{gray!30} 46.63 \\
    \midrule
  &AT w/ $\epsilon=\frac{0}{255}$&\cellcolor{gray!20}75.42 & 32.27 & 5.20 & 0.27 & 0.01&\cellcolor{gray!20}78.38 & 26.28 & 3.31 & 0.11 & 0.00\\
  &AT w/ $\epsilon=\frac{1}{255}$&69.83 &\cellcolor{gray!20} 57.94 & 44.98 & 24.15 & 5.05&74.54 &\cellcolor{gray!20} 61.34 & 47.99 & 23.62 & 4.41\\
  \textbf{CIFAR100}&AT w/ $\epsilon=\frac{2}{255}$&66.11 & 57.53 & \cellcolor{gray!20}48.79 & 31.39 & 10.76&71.43 & 61.99 & \cellcolor{gray!20} 51.73 & 33.02 & 10.85\\
  &AT w/ $\epsilon=\frac{4}{255}$&61.08 & 54.08 & 47.07 &\cellcolor{gray!20} 34.30 & 15.93&67.18 & 60.18 & 53.32 & \cellcolor{gray!20} 39.41 & 18.32\\
  &AT w/ $\epsilon=\frac{8}{255}$&55.62 & 50.18 & 44.94 & 35.20 & \cellcolor{gray!20}20.10&60.24 & 54.80 & 49.09 & 39.24 &\cellcolor{gray!20}  22.95\\
  &TCN (Ours)&\cellcolor{gray!30}70.71 &\cellcolor{gray!30} 55.91 &\cellcolor{gray!30} 47.01 &\cellcolor{gray!30} 38.86 &\cellcolor{gray!30} 20.34&\cellcolor{gray!30}75.42 &\cellcolor{gray!30} 62.73 &\cellcolor{gray!30} 52.12 &\cellcolor{gray!30} 37.76 &\cellcolor{gray!30} 22.21 \\\midrule
  &AT w/ $\epsilon=\frac{0}{255}$&\cellcolor{gray!20}60.40 & 35.29 & 5.03 & 0.03 & 0.00&\cellcolor{gray!20}69.84 & 19.75 & 0.55 & 0.02 & 0.01\\
  &AT w/ $\epsilon=\frac{1}{255}$&60.23 &\cellcolor{gray!20} 52.95 & 47.03 & 36.28 & 7.23&69.08 &\cellcolor{gray!20} 59.79 & 51.29 & 32.07 & 3.47\\
  \textbf{Tiny-ImageNet} &AT w/ $\epsilon=\frac{2}{255}$&58.17 & 52.67 &\cellcolor{gray!20} 48.21 & 41.35 & 21.45&67.50 & 60.56 &\cellcolor{gray!20} 54.41 & 44.75 & 15.27\\
  &AT w/ $\epsilon=\frac{4}{255}$&55.44 & 51.36 & 47.94 &\cellcolor{gray!20} 42.58 & 33.46&64.43 & 59.26 & 54.76 &\cellcolor{gray!20} 47.58 & 33.59\\
  &AT w/ $\epsilon=\frac{8}{255}$&51.50 & 48.24 & 45.77 & 41.87 &\cellcolor{gray!20} 36.76&59.82 & 56.05 & 52.45 & 47.14 &\cellcolor{gray!20}  39.61 \\
  &TCN (Ours)&\cellcolor{gray!30}54.99 & \cellcolor{gray!30}50.21 &\cellcolor{gray!30} 47.14 &\cellcolor{gray!30} 42.34 &\cellcolor{gray!30} 34.01&\cellcolor{gray!30} 68.48 & \cellcolor{gray!30} 60.77 &\cellcolor{gray!30} 54.03 &\cellcolor{gray!30} 42.02 &\cellcolor{gray!30} 21.82\\
    \bottomrule
    \end{tabular}
}

    \label{tab:acc_across_budget}
\end{table}

We evaluate TCN across three datasets (CIFAR-10, CIFAR-100, and Tiny-ImageNet) 
and two backbone architectures (ResNet-18 and WRN-28-10), with results summarized 
in Table~\ref{tab:acc_across_budget}. We consider the following set of threat 
levels: $\mathcal{E} = \{\frac{0}{255}, \frac{1}{255}, \frac{2}{255}, 
\frac{4}{255}, \frac{8}{255}\}$. For each vanilla adversarial training (AT) 
baseline, a single fixed budget $\epsilon \in \mathcal{E}$ is used for training. 
For TCN, we instead apply distributional adversarial training over the full set 
$\mathcal{E}$. All models are evaluated at every budget in $\mathcal{E}$.
From these results, we draw the following observations:
(1) Performance conflict in separately trained models.
    Models trained at different perturbation budgets are highly specialized: each performs best on its matched test budget (diagonal entries), but degrades significantly under mismatched settings. This reveals poor cross-budget generalization.
(2) TCN approximates the optimal diagonal.
    TCN achieves performance consistently close to the best diagonal entries using a single conditional model, effectively recovering multiple specialized behaviors within one unified framework.

\subsection{State-of-the-art Performance}

\begin{table}[h!]
    \centering
    \vspace{-0.15in}
\caption{Comparison of accuracy (\%) on CIFAR-10 with ResNet-18. We compare natural training, fixed-budget and uniform-budget PGD-AT, 
representative single-model adversarial training methods (TRADES, MART, NuAT), and 
multi-model baselines (MoRE, Ensemble). Our TCN  outperforms all baselines on average performance.}
\vspace{0.1in}
    \begin{tabular}{c|ccccc|c|cc}
\toprule
\textbf{Method} $\backslash$ \textbf{Budget}&  $\mathbf{\frac{0}{255}}$&$\mathbf{\frac{1}{255}}$&$\mathbf{\frac{2}{255}}$&$\mathbf{\frac{4}{255}}$&$\mathbf{\frac{8}{255}}$&\textbf{Avg.}&\textbf{\# Param.}
\\\midrule
  Natural Training& \textbf{95.09} & 57.86 & 8.94 & 0.14 & 0.00&32.41&11.17M\\
  \midrule
  PGD-AT w/ $\epsilon=\frac{8}{255}$&84.07 & 80.27 & 75.82 & 65.71 &  43.10&69.79&11.17M\\
  PGD-AT w/ Uniform $\epsilon$&87.59 & 83.78 & 78.58 & 67.01 & 39.21&71.23&11.17M\\
  \midrule
  TRADES ($\lambda =2.0$)&81.92&78.00&73.51&62.89&41.50&67.56&11.17M\\
  TRADES ($\lambda =0.2$)&91.49&62.74& 27.58&10.27&3.31&39.08&11.17M\\
  MART ($\lambda =5.0$)&80.92&77.36&73.46&64.57&44.85&68.23&11.17M\\
  MART ($\lambda =0.5$)&82.29&78.18&73.66&63.56&42.25&67.99&11.17M\\
  NuAT&80.23&77.61& 75.1&69.13&\textbf{46.18}&69.65&11.17M\\
  \midrule
  MoRE&84.52&77.10&72.80&63.10&41.22&67.75&55.87M\\
  Ensemble&94.46& 87.27&73.95&49.89&20.13&65.14&55.87M\\
  \midrule
  TCN (Fixed Cond.)&87.44 & 83.73 & 79.58 & \textbf{69.80}  & 45.16&73.14&11.69M \\
  TCN (Adaptive Cond.) &93.85 & \textbf{87.46} &  \textbf{80.49} & 66.36 &  45.16&\textbf{74.66}&11.69M \\
\bottomrule

    \end{tabular}

\vspace{-0.15in}
    \label{tab:resnet18_cifar10_sota}
\end{table}

To provide a comprehensive comparison, we consider four groups of baselines:
(i) vanilla natural training,
(ii) PGD-based adversarial training with fixed perturbation budget and uniform budget,
(iii) representative adversarial training methods such as TRADES, MART, and NuAT, and
(iv) multi-model baselines such as MoRE and ensemble models. For TCN, we consider two variants: TCN (Fixed Cond.), which fixes the 
conditioning threat level at $\epsilon = \frac{8}{255}$ regardless of the test 
budget, and TCN (Adaptive Cond.), which sets the conditioning threat 
level to match the test budget.
All methods are evaluated on CIFAR-10 with ResNet-18.
We make the following observations from Table~\ref{tab:resnet18_cifar10_sota}:
    (1) PGD-based adversarial training is budget-specialized.
    Fixed-budget PGD-AT ($\epsilon{=}\frac{8}{255}$) and uniform-budget PGD-AT achieve
    69.79\% and 71.23\% average accuracy, respectively, but neither generalizes
    consistently across mismatched test budgets.
   (2) Advanced single-model methods remain limited.
    TRADES, MART, and NuAT offer varying robustness--accuracy trade-offs but fall short
    on cross-budget generalization. 
    (3) Multi-model baselines cost more but gain little.
    MoRE and Ensemble use 55.87M parameters ($5\times$ more than single-model methods)
    yet achieve only 67.75\% and 65.14\% average accuracy, showing that naively
    combining models does not address cross-budget generalization.
    (4) TCN achieves the best performance with a single model.
    TCN (adaptive condition) reaches 74.66\% average accuracy with only 11.69M
    parameters, outperforming all baselines including multi-model methods. TCN (fixed
    condition) also achieves 73.14\%, already surpassing every competitor, confirming
    TCN as a scalable and parameter-efficient solution for adaptive adversarial
    robustness.

\subsection{Transferability Analysis}

To further evaluate the generalization ability of TCN across perturbation levels, we study its \emph{transferability} under mismatched conditioning. 
In particular, we consider a realistic setting where the true threat level is unknown at test time, and analyze how the model performs when conditioned on a fixed perturbation level. 
We conduct two complementary analyses, including performance comparison and representation visualization.

\begin{wraptable}{r}{0.5\textwidth}
    \centering
    \vspace{-0.01in}
    \caption{Transfer performance under mismatched threat levels on CIFAR-10 with ResNet-18. 
TCN conditioned on $\epsilon_i$ outperforms the AT model separately trained on $\epsilon_i$ 
in every row-wise comparison. 
}
\renewcommand{\arraystretch}{1.1}
\setlength{\tabcolsep}{3pt} 
\resizebox{0.5\textwidth}{!}{%

    \begin{tabular}{c|c|ccccc|c}

\toprule
\multicolumn{2}{c|}{\textbf{Model} $\backslash$ \textbf{Budget}} & $\mathbf{\epsilon_0}$&$\mathbf{\epsilon_1}$&$\mathbf{\epsilon_2}$&$\mathbf{\epsilon_3}$&$\mathbf{\epsilon_4}$& \textbf{Avg.}
\\\midrule
\multirow{5}{*}{\textbf{AT}}
  & $\gF_{\theta^*_0}(\cdot)$ &95.09 & 57.86 & 8.94 & 0.14 & 0.00 & 32.41\\
  & $\gF_{\theta^*_1}(\cdot)$&93.09 & 86.80 & 76.80 & 51.01 & 20.50 & 65.64\\
  & $\gF_{\theta^*_2}(\cdot)$&91.11 & 85.81 & 79.28 & 61.39 & 28.68 & 69.25\\
  & $\gF_{\theta^*_3}(\cdot)$&88.51 & 83.92 & 78.61 & 65.51 & 36.57 & 70.62\\
  & $\gF_{\theta^*_4}(\cdot)$&84.07 & 80.27 & 75.82 & 65.71 & 43.10 & 69.79\\
  \midrule
\multirow{5}{*}{\textbf{TCN}}
  & $\gT_{\epsilon_0}\circ\gF_{\theta^*}(\cdot)$&93.85 & 86.60 & 74.50 & 44.93 & 11.68 & 62.31\\
  & $\gT_{\epsilon_1}\circ\gF_{\theta^*}(\cdot)$&92.86 & 87.46 & 79.48 & 58.06 & 22.16 & 68.00\\
  & $\gT_{\epsilon_2}\circ\gF_{\theta^*}(\cdot)$&92.26 & 87.47 & 80.49 & 62.26 & 26.57 & 69.81\\
  & $\gT_{\epsilon_3}\circ\gF_{\theta^*}(\cdot)$&90.96 & 86.80 & 81.03 & 66.36 & 33.91 & 71.81\\
  & $\gT_{\epsilon_4}\circ\gF_{\theta^*}(\cdot)$&87.44 & 83.73 & 79.58 & 69.80 & 45.16 & 73.14\\
    \bottomrule
    \end{tabular}
}

    \vspace{-0.15in}
    \label{tab:resnet18_cifar10_transfer}
\end{wraptable}

\textbf{Performance Transferability.} 
In practice, the adversarial threat level is often unknown or dynamically changing at test time.
It is therefore important to evaluate how well models transfer under mismatched perturbation budgets.
Vanilla AT trains a separate model $\mathcal{F}_{\theta^*_i}$ for each budget $\epsilon_i$, while TCN uses a single backbone $\mathcal{F}_{\theta^*}$ trained with a uniform distribution over budgets, adapting via the conditioning operator $\mathcal{T}_{\epsilon_i}$. For a fair comparison, we evaluate each AT specialist against the TCN variant conditioned on the same budget. As shown in Table~\ref{tab:resnet18_cifar10_transfer}, TCN achieves higher average accuracy in every row, demonstrating consistently superior cross-budget transferability. This stems from learning a shared representation across perturbation levels, which enforces smoothness across $\epsilon$ and preserves strong performance even under mismatched test conditions where independently trained AT models degrade substantially.

\begin{figure}[h!]
    \centering
    \includegraphics[width=0.99\linewidth]{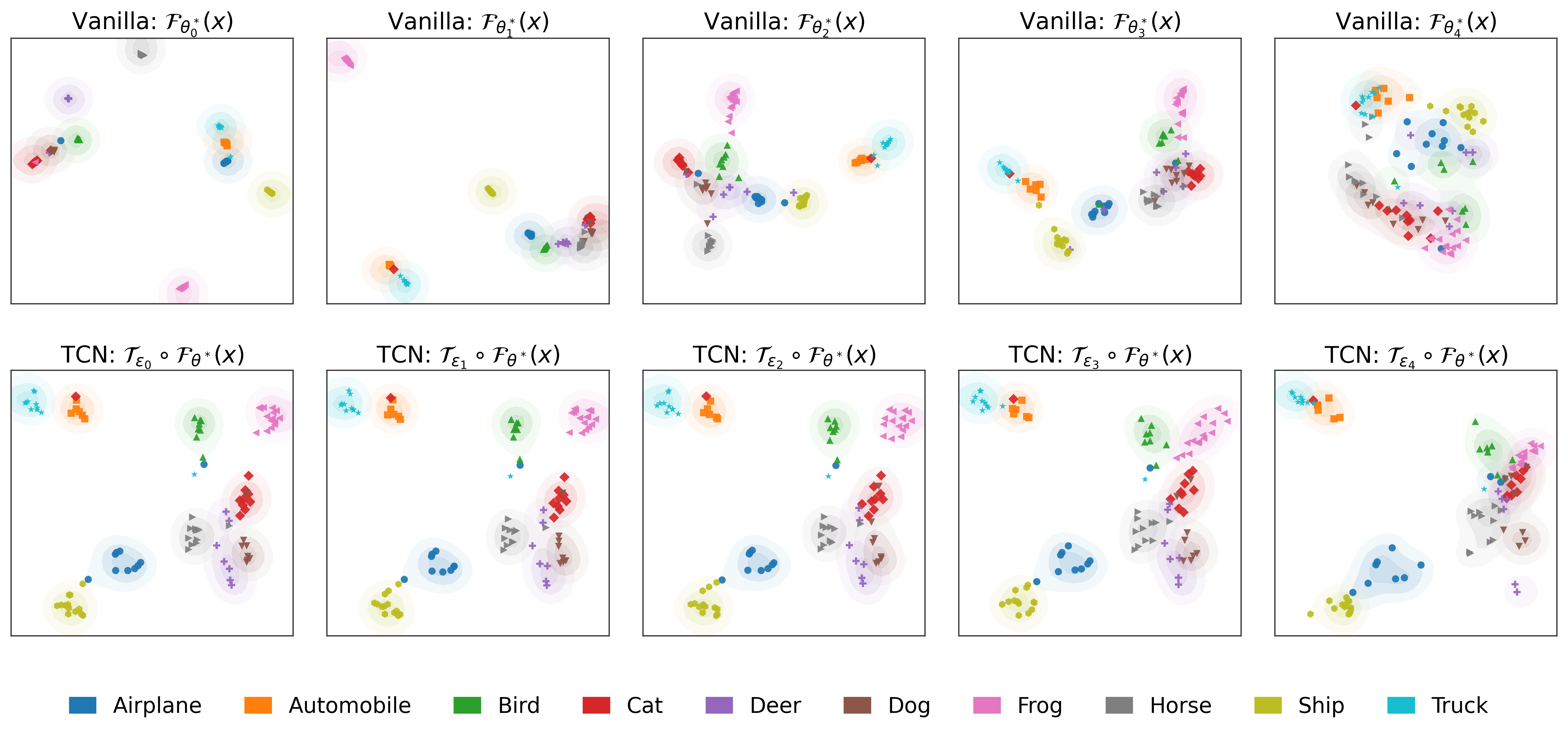}
\caption{
UMAP visualization of learned representations under different adversarial levels.
\textbf{Top row}: independently trained vanilla models at different noise levels.
\textbf{Bottom row}: TCN conditioned on different adversarial levels.
Under mismatched conditions, vanilla representations degrade and fragment, whereas TCN preserves a coherent global structure, demonstrating stronger transferability.
}
    \label{fig:umap_visualization}
\vspace{-0.15in}
\end{figure}

\textbf{Visualization.}
Fig.\ref{fig:umap_visualization} visualizes UMAP embeddings of the final hidden representations under mismatched adversarial conditions, using the same setup as ``Performance Transferability.'' Under matched conditions, both vanilla adversarial training and TCN produce well-separated class clusters. Under mismatched conditions, however, vanilla representations progressively fragment and lose discriminability, whereas TCN preserves a coherent global structure across adversarial levels. This indicates that TCN learns a family of representations parameterized by adversarial level, explaining its stronger transferability. Embedding difference analysis is provided in Appendix\ref{sec:embed_analysis}.

\subsection{Ablation Study}


\begin{figure}[h!]
    \centering
    \vspace{-0.15in}
    \includegraphics[width=0.99\linewidth]{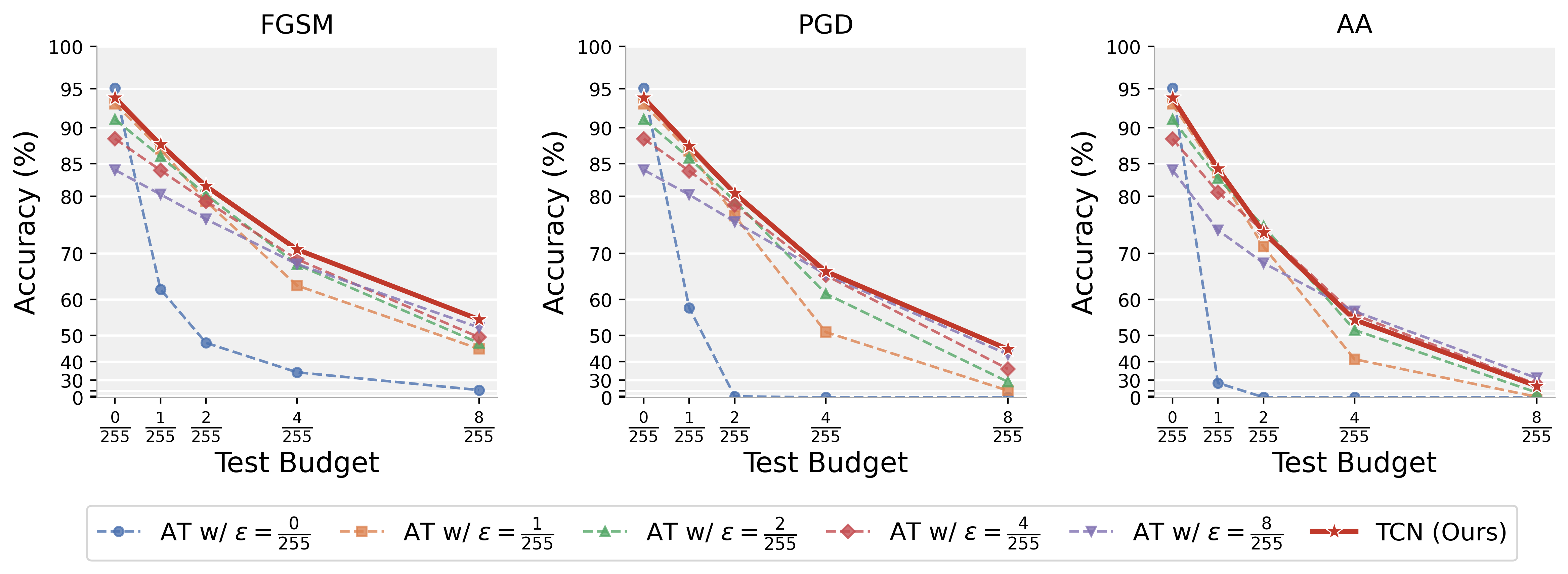}
    \caption{Robustness against FGSM, PGD, and AutoAttack (AA) 
    across varying test budgets. 
    }
    \label{fig:diff_threat}
    \vspace{-0.2in}
\end{figure}

\textbf{Different Threats.} We compare TCN against adversarially trained 
(AT) models, each specialized for a fixed training budget $\epsilon \in 
\{0, \frac{1}{255}, \frac{2}{255}, \frac{4}{255}, \frac{8}{255}\}$, under 
three attack types (FGSM, PGD, and AutoAttack) across varying test budgets 
(Figure~\ref{fig:diff_threat}). The experiments are performed on CIFAR-10 with 
ResNet18. As shown, AT models tend to overfit to 
their specific training budget, performing well near their trained 
$\epsilon$ but degrading significantly at other test budgets. In 
contrast, TCN consistently matches or surpasses the best-performing AT 
baseline across all test budgets with one single model. Additional experimental results are provided in Appendix~\ref{sec:diff_threat_app}.


\begin{wraptable}{r}{0.42\textwidth}
    \centering
    \vspace{-0.0in}
    \caption{Layer-wise $\Gamma(\epsilon)$ values ($\times 10^{-2}$). Larger noise leads to smaller $\Gamma(\epsilon)$, indicating stronger feature suppression.}
    \setlength{\tabcolsep}{2pt}
    \renewcommand{\arraystretch}{0.9}
    \vspace{-0.1in}
    \resizebox{0.42\textwidth}{!}{%
    \begin{tabular}{l|ccccc}
\toprule
\textbf{Layer}  $\boldsymbol{\ell}$  & $\boldsymbol{\epsilon=\frac{0}{255}}$ & $\boldsymbol{\epsilon=\frac{1}{255}}$ & $\boldsymbol{\epsilon=\frac{2}{255}}$ & $\boldsymbol{\epsilon=\frac{4}{255}}$ & $\boldsymbol{\epsilon=\frac{8}{255}}$ \\
\midrule
$\ell=$ 1  & 3.17  & 2.57  & 2.08  & 1.25  & -0.05 \\
$\ell=$ 2  & 2.50  & 1.85  & 1.31  & 0.37  & -1.10 \\
$\ell=$ 3  & 4.28  & 3.94  & 3.61  & 2.98  & 1.98 \\
$\ell=$ 4  & 7.07  & 6.39  & 5.86  & 4.97  & 3.57 \\
$\ell=$ 5  & 2.80  & 2.66  & 2.50  & 2.17  & 1.64 \\
$\ell=$ 6  & 5.02  & 4.79  & 4.55  & 4.10  & 3.39 \\
$\ell=$ 7  & 2.00  & 1.66  & 1.37  & 0.87  & 0.09 \\
$\ell=$ 8  & 3.55  & 3.05  & 2.65  & 2.00  & 0.97 \\
$\ell=$ 9  & 2.96  & 2.72  & 2.50  & 2.10  & 1.48 \\
$\ell=$ 10 & 4.48  & 4.33  & 4.19  & 3.96  & 3.58 \\
$\ell=$ 11 & 1.72  & 1.37  & 1.02  & 0.36  & -0.69 \\
$\ell=$ 12 & 3.99  & 3.61  & 3.26  & 2.63  & 1.63 \\
$\ell=$ 13 & 1.15  & 1.04  & 0.93  & 0.72  & 0.40 \\
$\ell=$ 14 & 4.65  & 4.56  & 4.49  & 4.36  & 4.17 \\
$\ell=$ 15 & 0.88  & 0.73  & 0.60  & 0.37  & 0.02 \\
$\ell=$ 16 & 7.67  & 7.41  & 7.22  & 6.90  & 6.40 \\
\bottomrule
    \end{tabular}}

    \vspace{-0.1in}
    \label{tab:gamma_analysis}
\end{wraptable}

\textbf{Behavior analysis of $\Gamma(\epsilon)$.}
In TCN, we use the channel-wise conditional modulation $
\mathcal{T}_\epsilon\circ\mathcal{F} (\mathbf{x}) = (1+\Gamma(\epsilon)) \odot \mathcal{F}(\mathbf{x}) + \mathrm{B}(\epsilon)$,
where the noise-conditioned parameters $\Gamma(\epsilon)$ and $\mathrm{B}(\epsilon)$ adaptively adjust the embedding $h$. In particular, $\Gamma(\epsilon)$ controls the rescaling of the embedding. As reported in Table~\ref{tab:gamma_analysis}, the magnitude of $\Gamma(\epsilon)$ tends to increase as the noise level $\epsilon$ becomes larger. This behavior suggests that under stronger perturbations, TCN applies a stronger modulation to the feature representation. More specifically, the resulting factor $(1+\Gamma(\epsilon))$ tends to impose a smaller effective scaling on the embedding, which suppresses unstable feature responses and reduces the influence of perturbations in the representation space. This observation indicates that TCN adaptively contracts the embedding under higher noise levels, helping improve feature stability and robustness.

\begin{wraptable}{r}{0.42\textwidth}
    \centering
    \vspace{-0.1in}
\caption{Performance on AG News under TextFooler attacks with varying similarity thresholds $\epsilon$. TCN achieves more balanced clean-robust performance across perturbation levels than fixed-level AT.}
    \setlength{\tabcolsep}{3pt}
    \renewcommand{\arraystretch}{1.0}
    \resizebox{0.42\textwidth}{!}{%
    \begin{tabular}{cccccccccccc}
    \toprule
      Method $\backslash$ Threshold $\epsilon$  & 0.0&0.1&0.3&0.5&0.7&0.9 \\
      \midrule
       AT, $\epsilon=0.0$ &94.2&82.0&40.0&17.3& 8.4& 7.9 \\
        AT, $\epsilon=0.1$&94.1 & 82.5 & 39.3 & 16.3 & 7.8 & 7.0\\
       AT, $\epsilon=0.3$&94.5 & 87.0 & 45.9 & 19.4 & 8.4 & 8.2\\
       AT, $\epsilon=0.5$&94.3 & 87.1 & 47.9 & 29.6 & 15.7 & 14.7\\
       AT, $\epsilon=0.7$&93.4 & 85.3 & 48.3 & 28.4 & 17.5 & 17.2\\
        AT, $\epsilon=0.9$&93.1 & 83.9 & 50.2 & 33.9 & 23.5 & 23.3\\
        \midrule
       TCN &94.1 & 86.0 & 53.1 & 31.5 & 20.4 & 19.8 \\
    \bottomrule
    \end{tabular}
    }

    \label{tab:agnews_textfooler}
    \vspace{-0.1in}
\end{wraptable}

\textbf{Language modeling.} 
Beyond visual classification, we further validate our method in the language domain on AG News using a BERT-based classifier under TextFooler attacks with different tolerated word embedding difference thresholds $\epsilon$. As in the vision setting, fixed-level adversarial training exhibits a clear trade-off: models trained with weaker perturbations preserve higher clean accuracy but perform poorly under stronger attacks, while models trained with stronger perturbations improve robustness at the cost of clean performance. In contrast, our conditional method achieves stronger overall performance across different similarity levels, showing that the proposed framework generalizes beyond image classification. Notably, even when conditioned on a single fixed level, our model still maintains competitive performance across a wide range of perturbation strengths, further demonstrating its transferability and supporting the view that robustness is better modeled as a continuous spectrum rather than a collection of isolated training objectives.

\begin{table}[h!]
    \centering
\vspace{-0.15in}
    \caption{Ablation study on threat-level embeddings. }
\resizebox{0.9\textwidth}{!}{%
    \begin{tabular}{c|ccc|ccccccccc}
\toprule
&Discrete&Continuous&Fourier&  \textbf{0.0} &\textbf{0.05}&\textbf{0.1}&\textbf{0.15}&\textbf{0.2}&\textbf{0.25}&\textbf{0.3}
\\
\midrule
&\checkmark&&&82.5 & 75.1 & 63.2 & 48.5 & 33.0 & 18.3 & 5.8 \\
TCN&&\checkmark&&86.4 & 76.6 & 62.7 & 44.0 & 28.1 & 14.2 & 3.5 \\
&&\checkmark&\checkmark&90.4 & 77.4 & 65.4 & 51.7 & 38.5 & 25.8 & 14.1\\
\midrule
 &\multicolumn{3}{c|}{AT w/ $\epsilon=0.0$}&91.2 & 49.9 & 7.4 & 0.3 & 0.0 & 0.0 & 0.0 \\
Baseline &\multicolumn{3}{c|}{ AT w/ $\epsilon=0.3$}&79.1 & 70.4 & 60.0 & 48.5 & 36.2 & 23.8 & 12.2\\ 
&\multicolumn{3}{c|}{AT w/ Uniform $\epsilon$}&84.4 & 75.2 & 63.2 & 47.8 & 31.8 & 16.3 & 3.6\\
    \bottomrule

    \end{tabular}
}
\vspace{-0.12in}
    \label{tab:ablation_noise_embed}
\end{table}

\textbf{Threat-level embedding.}
Table~\ref{tab:ablation_noise_embed} compares three conditioning strategies: discrete embeddings (\texttt{nn.Embedding}), raw continuous scalar input, and continuous Fourier-based embeddings (random Fourier features). Discrete embeddings lack smooth interpolation across noise levels, while raw continuous inputs provide insufficient expressiveness, and both fail to achieve consistently strong performance across budgets. Fourier-based embeddings resolve this by mapping scalar noise levels into a richer feature space, enabling smoother and more structured conditioning that yields the best overall performance across all perturbation regimes.

\begin{wraptable}{r}{0.42\textwidth}
    \centering
    \vspace{-0.15in}
\caption{Robustness on unseen perturbation budgets. TCN conditions on the nearest in-distribution budget and consistently outperforms all baselines.}
\vspace{-0.1in}
\resizebox{0.42\textwidth}{!}{%

    \begin{tabular}{c|cccc}
\toprule
\textbf{Method} $\backslash$ \textbf{Budget}& $\mathbf{\frac{3}{255}}$&$\mathbf{\frac{5}{255}}$&$\mathbf{\frac{6}{255}}$&$\mathbf{\frac{7}{255}}$
\\
\midrule
  AT w/ $\epsilon=\frac{0}{255}$&1.15&0.01&0.0&0.0\\
  AT w/ $\epsilon=\frac{1}{255}$&64.02&39.67&31.23&24.93\\
  AT w/ $\epsilon=\frac{2}{255}$&70.98&51.46&42.65&34.67\\
  AT w/ $\epsilon=\frac{4}{255}$&72.45&58.19&50.6&43.32\\
 AT w/ $\epsilon=\frac{8}{255}$&70.99&60.05&54.47&48.70\\
 \midrule
  TCN&74.43&64.04&57.63&51.35\\
    \bottomrule
    \end{tabular}
}

\vspace{-0.2in}
\label{tab:resnet18_cifar10_ood}
\end{wraptable}

\textbf{Generalization to Unseen Perturbation Budgets.}
Beyond the training budgets, we evaluate model generalization to unseen (out-of-distribution) perturbation levels. Specifically, we treat $\{\frac{0}{255}, \frac{1}{255}, \frac{2}{255}, \frac{4}{255}, \frac{8}{255}\}$ as in-distribution budgets and $\{\frac{3}{255}, \frac{5}{255}, \frac{6}{255}, \frac{7}{255}\}$ as out-of-distribution budgets. For TCN, the conditioning variable is set to the nearest upper in-distribution budget. 
As shown in Table~\ref{tab:resnet18_cifar10_ood}, TCN consistently outperforms all baselines across all unseen budgets. In contrast, standard adversarial training models exhibit strong sensitivity to their training budget and degrade when evaluated at mismatched levels. These results demonstrate that TCN generalizes effectively across unseen perturbation strengths, maintaining robust performance beyond the training distribution.

\begin{wraptable}{r}{0.45\textwidth}
\centering
\vspace{-0.12in}
\setlength{\tabcolsep}{2pt} 
\caption{Computational complexity of ResNet18 and TCN-ResNet18 on CIFAR-10.
}
\label{tab:complexity}
\resizebox{0.42\textwidth}{!}{%
\begin{tabular}{cccc}
\toprule
Model & \#Params (M) & Model Size (MB) & FLOPs (G) \\
\midrule
ResNet18         & 11.174 & 42.66 & 1.1158 \\
TCN-ResNet18     & 11.690 & 44.63 & 1.1168 \\
\midrule
$\Delta$         & +4.6\% &+4.6\% & +0.09\% \\
\bottomrule
\end{tabular}
}
\end{wraptable}
\textbf{Complexity Analysis.}
Table~\ref{tab:complexity} compares the computational footprint of TCN-ResNet18 against vanilla ResNet18. The noise-conditional modules add only $0.52$M parameters ($+4.6\%$), translating to a $1.97$\,MB increase in model size. Because FiLM applies per-channel affine modulations rather than dense matrix multiplications, the forward FLOPs of TCN-ResNet18 ($1.1168$\,G) are virtually identical to those of ResNet18 ($1.1158$\,G), a relative increase of less than $0.1\%$. Overall, TCN-ResNet18 introduces noise-level conditioning at essentially no additional inference cost, making it a practical drop-in replacement for ResNet18.

\vspace{-0.1in}
\section{Related Works}
\vspace{-0.1in}



Several lines of work have sought to train a model capable of defending 
against diverse adversarial threats. One direction exposes the model to 
multiple threat types during training~\citep{tramer2019adversarial, 
maini2020adversarial, cai2018curriculum}, enabling it to handle heterogeneous 
attacks simultaneously. Another direction redesigns the training objective: 
methods such as TRADES~\citep{zhang2019theoretically} and 
MART~\citep{wang2019improving} introduce new loss formulations to better 
balance accuracy and robustness. 
A complementary line of work pursues \emph{universal} robust architectures, whose
robustness is built into the model itself rather than tailored to a particular
attack, and therefore transfers across heterogeneous
threats~\citep{hou2024protransformer,hou2024robust,hou2025robustness,hou2025boosting}.
Scaling model capacity has also shown 
promise, with larger models exhibiting improved robustness across 
budgets~\citep{rice2020overfitting}. Ensemble-based 
methods~\citep{pang2019improving, strauss2017ensemble, cheng2021mixture} 
have similarly been explored to improve cross-budget generalization. Despite 
these efforts, performance gains remain limited across all these approaches: 
redesigned objectives offer marginal improvement over vanilla AT, while 
scaling model capacity and ensemble methods both suffer from poor 
computational scalability, making them impractical for real-world deployment.

\vspace{-0.1in}
\section{Conclusion}
\vspace{-0.1in}
\label{sec:conclusion}

We study adversarial robustness across diverse threat levels and show that conventional adversarial training generalizes poorly across perturbation regimes due to its budget-specific formulation. We reformulate robustness as a threat-conditional prediction problem and propose TCN, a single model that adapts continuously to the perturbation level by learning a shared representation across adversarial scales. Extensive experiments demonstrate that TCN achieves strong and consistent performance across a wide range of threat levels, generalizes to unseen budgets, and remains parameter-efficient. Our results suggest that modeling adversarial robustness as a continuous conditional learning task provides a more scalable and principled solution for dynamic threat environments.
While TCN transfers well across threat levels, it still requires the threat level
as an input. Handling an unknown threat level at inference via Bayesian
inference or adversarial detection is left to future work.

\newpage

\bibliography{example_paper}
\bibliographystyle{icml2026}


\appendix
\newpage
\section{Additional Experiments}

\subsection{Experimental Results under Different Attacks}
\label{sec:diff_threat_app}

Table~\ref{tab:resnet18_cifar10_diff_threat} and 
Table~\ref{tab:wideresnet2810_cifar10_diff_threat} report accuracy under 
FGSM, PGD, and AutoAttack across varying test budgets on CIFAR-10 with 
ResNet18 and WideResNet-28-10, respectively. AT models show clear 
overfitting to their training budget: each AT model performs best at its 
matched test budget (highlighted cells on the diagonal) but degrades 
noticeably outside that range, with the effect being most severe under 
stronger attacks such as PGD and AutoAttack---AT w/ $\epsilon=\frac{0}{255}$ 
drops to $0.00\%$ at test budget $\frac{8}{255}$ under both attacks. In 
contrast, TCN consistently matches or surpasses the best-performing AT 
baseline at every test budget across all three attack types and both 
architectures, demonstrating that TCN generalizes robustly across diverse 
threat levels without being tied to any fixed training budget.

\begin{table}[h!]
    \centering
    \caption{Different threats (CIFAR10, ResNet18).}
    \vspace{0.1in}
    \begin{tabular}{c|ccccc}
\toprule
\textbf{Method} $\backslash$ \textbf{Budget}& $\mathbf{\frac{0}{255}}$&$\mathbf{\frac{1}{255}}$&$\mathbf{\frac{2}{255}}$&$\mathbf{\frac{4}{255}}$&$\mathbf{\frac{8}{255}}$
\\
\midrule
\multicolumn{6}{c}{FGSM}\\
AT w/ $\epsilon=\frac{0}{255}$&\cellcolor{gray!20} 95.09&62.42 & 47.46 & 34.76 & 21.11\\
AT w/ $\epsilon=\frac{1}{255}$&93.09& \cellcolor{gray!20}87.09 & 79.20 & 63.28 & 45.27\\
AT w/ $\epsilon=\frac{2}{255}$&91.11& 86.03 & \cellcolor{gray!20}80.29 & 67.79 & 47.37\\
AT w/ $\epsilon=\frac{4}{255}$&88.51& 84.02 & 79.24 & \cellcolor{gray!20}68.86 & 49.35\\
AT w/ $\epsilon=\frac{8}{255}$&84.07& 80.33 & 76.25 & 67.89 & \cellcolor{gray!20}52.39\\
TCN (Ours)&\cellcolor{gray!30}93.85&\cellcolor{gray!30}87.72&\cellcolor{gray!30}81.62&\cellcolor{gray!30}70.77&\cellcolor{gray!30}54.74 \\
\midrule
\multicolumn{6}{c}{PGD}\\
AT w/ $\epsilon=\frac{0}{255}$&\cellcolor{gray!20} 95.09 & 57.86 & 8.94 & 0.14 & 0.00\\
AT w/ $\epsilon=\frac{1}{255}$&93.09 & \cellcolor{gray!20}86.80 & 76.80 & 51.01 & 20.50\\
AT w/ $\epsilon=\frac{2}{255}$&91.11 & 85.81 & \cellcolor{gray!20} 79.28 & 61.39 & 28.68\\
AT w/ $\epsilon=\frac{4}{255}$&88.51 & 83.92 & 78.61 & \cellcolor{gray!20} 65.51 & 36.57\\
AT w/ $\epsilon=\frac{8}{255}$&84.07 & 80.27 & 75.82 & 65.71 & \cellcolor{gray!20} 43.10\\
TCN (Ours)&\cellcolor{gray!30}93.85 & \cellcolor{gray!30} 87.46 & \cellcolor{gray!30} 80.49 & \cellcolor{gray!30} 66.36 & \cellcolor{gray!30} 45.16 \\
\midrule
\multicolumn{6}{c}{AA}\\
AT w/ $\epsilon=\frac{0}{255}$&\cellcolor{gray!20} 95.09&27.78 & 2.21 & 0.10 & 0.00\\
AT w/ $\epsilon=\frac{1}{255}$&93.09&\cellcolor{gray!20}83.65 & 71.31 & 41.12 & 7.12\\
AT w/ $\epsilon=\frac{2}{255}$&91.11&82.85 & \cellcolor{gray!20}75.03 & 51.65 & 17.95\\
AT w/ $\epsilon=\frac{4}{255}$&88.51&80.64 & 74.22 & \cellcolor{gray!20}56.07 & 26.78\\
AT w/ $\epsilon=\frac{8}{255}$&84.07&74.32 & 68.10 & 56.97 & \cellcolor{gray!20}31.29\\
TCN (Ours)&\cellcolor{gray!30}93.85&\cellcolor{gray!30}84.25&\cellcolor{gray!30}73.92&\cellcolor{gray!30}54.66&\cellcolor{gray!30}25.28 \\
    \bottomrule
    \end{tabular}
    \label{tab:resnet18_cifar10_diff_threat}
\end{table}

\begin{table}[h!]
    \centering
    \caption{Different threats (CIFAR10, WideResNet-28-10).}
    \vspace{0.1in}

    \begin{tabular}{c|ccccc}
\toprule
\textbf{Method} $\backslash$ \textbf{Budget}& $\mathbf{\frac{0}{255}}$&$\mathbf{\frac{1}{255}}$&$\mathbf{\frac{2}{255}}$&$\mathbf{\frac{4}{255}}$&$\mathbf{\frac{8}{255}}$
\\
\midrule
\multicolumn{6}{c}{FGSM}\\
AT w/ $\epsilon=\frac{0}{255}$&\cellcolor{gray!20}95.37 & 56.63 & 35.55 & 19.69 & 8.42\\
AT w/ $\epsilon=\frac{1}{255}$&94.67 & \cellcolor{gray!20}88.84 & 80.90 & 65.06 & 45.22\\
AT w/ $\epsilon=\frac{2}{255}$&93.22 & 88.69 & \cellcolor{gray!20}82.93 & 70.21 & 50.45\\
AT w/ $\epsilon=\frac{4}{255}$&90.91 & 86.80 & 82.54 & \cellcolor{gray!20}72.26 & 53.72\\
AT w/ $\epsilon=\frac{8}{255}$&86.40 & 82.66 & 78.59 & 70.25 & \cellcolor{gray!20}54.06\\
TCN (Ours)&\cellcolor{gray!30}94.56&\cellcolor{gray!30} 88.42 & \cellcolor{gray!30}81.99 &\cellcolor{gray!30} 70.14 &\cellcolor{gray!30} 54.57\\
\midrule
\multicolumn{6}{c}{PGD}\\
  AT w/ $\epsilon=\frac{0}{255}$&\cellcolor{gray!20}95.37 & 50.41 & 5.44 & 0.00 & 0.00\\
  AT w/ $\epsilon=\frac{1}{255}$&94.67 & \cellcolor{gray!20}88.55 & 78.66 & 50.20 & 14.12\\
 AT w/ $\epsilon=\frac{2}{255}$&93.22 & 88.50 & \cellcolor{gray!20}81.91 & 63.93 & 28.29\\
  AT w/ $\epsilon=\frac{4}{255}$&90.91 & 86.72 & 81.90 & \cellcolor{gray!20} 68.66 & 39.79\\
  AT w/ $\epsilon=\frac{8}{255}$&86.40 & 82.57 & 78.13 & 68.15 & \cellcolor{gray!20} 45.09\\
  TCN (Ours)& \cellcolor{gray!30} 94.56 & \cellcolor{gray!30} 88.49 & \cellcolor{gray!30} 82.32 & \cellcolor{gray!30} 68.87 & \cellcolor{gray!30} 46.63 \\

\midrule

\multicolumn{6}{c}{AA}\\
  AT w/ $\epsilon=\frac{0}{255}$&\cellcolor{gray!20}95.37 &25.18 & 2.01 & 0.00 & 0.00\\
  AT w/ $\epsilon=\frac{1}{255}$&94.67 &\cellcolor{gray!20}85.66 & 73.62 & 44.23 & 7.42 \\
 AT w/ $\epsilon=\frac{2}{255}$&93.22 &85.56 & \cellcolor{gray!20}77.53 & 57.17 & 19.96 \\
  AT w/ $\epsilon=\frac{4}{255}$&90.91 &84.95 & 79.14 &\cellcolor{gray!20} 63.09 & 31.49 \\
  AT w/ $\epsilon=\frac{8}{255}$&86.40 &80.94 & 75.03 & 60.88 &\cellcolor{gray!20} 37.51 \\
  TCN (Ours)& \cellcolor{gray!30} 94.56 & \cellcolor{gray!30}86.36 & \cellcolor{gray!30} 76.03 & \cellcolor{gray!30} 52.76 & \cellcolor{gray!30} 24.47  \\
    \bottomrule
    \end{tabular}
\label{tab:wideresnet2810_cifar10_diff_threat}
\end{table}

\newpage
\subsection{Embedding Analysis}
\label{sec:embed_analysis}

To compare transferability across different adversarial levels between vanilla adversarial training and TCN, we quantify the differences in their learned representations under mismatched adversarial conditions. Specifically, we use a ResNet-18 backbone trained on CIFAR-10 and consider five adversarial noise levels:
$$\epsilon_0 = \frac{0}{255},
\epsilon_1 = \frac{1}{255},
\epsilon_2 = \frac{2}{255},
\epsilon_3 = \frac{4}{255},
\epsilon_4 = \frac{8}{255}.$$
For vanilla adversarial training, we independently train five models, one for each adversarial level, resulting in separate parameters $\{\theta_0^*, \theta_1^*, \theta_2^*, \theta_3^*, \theta_4^*\}$.
In contrast, TCN requires only a single model $\theta^*$ while allowing the adversarial level to be specified through conditioning.
We feed clean inputs ($\epsilon_0 = 0$) into all models and extract hidden representations at different depths. This yields two sets of embeddings:
$$\{\mathcal{F}_{\theta_0^*}(\mathbf{x}),\mathcal{F}_{\theta_1^*}(\mathbf{x}),\mathcal{F}_{\theta_2^*}(\mathbf{x}),\mathcal{F}_{\theta_3^*}(\mathbf{x}),\mathcal{F}_{\theta_4^*}(\mathbf{x})\}$$
for vanilla adversarial training, and
$$\{\mathcal{T}_{\epsilon_0}\circ\mathcal{F}_{\theta^*}(\mathbf{x}),\mathcal{T}_{\epsilon_1}\circ\mathcal{F}_{\theta^*}(\mathbf{x}),\mathcal{T}_{\epsilon_2}\circ\mathcal{F}_{\theta^*}(\mathbf{x}),\mathcal{T}_{\epsilon_3}\circ\mathcal{F}_{\theta^*}(\mathbf{x}),\mathcal{T}_{\epsilon_4}\circ\mathcal{F}_{\theta^*}(\mathbf{x})\}$$
for TCN.
We consider 7 representations: the input, the output of the input layer, 
the first convolutional layer, the four residual blocks, and the final 
logits from the classifier.
The results are summarized in Table~\ref{tab:embed_difference}. Overall, the representation differences induced by TCN across adversarial levels are substantially smaller than those produced by vanilla adversarial training. Moreover, in vanilla adversarial training, the independently trained models $\{\theta_i^*\}_{i=0}^4$ produce highly inconsistent embeddings, even when evaluated on the same clean inputs. By contrast, TCN exhibits a gradual increase in representation difference as the conditioning level $\epsilon_i$ increases, while the overall discrepancy remains much smaller than that of vanilla adversarial training. These results suggest that the noise-conditioned architecture of TCN learns a sequential family of related mappings, which regularize one another and improve transferability and generalization across adversarial levels.

\begin{table}[h!]
    \centering
    \caption{Embedding differences across layers under different adversarial levels. Vanilla adversarial training yields substantially larger representation discrepancies across separately trained models, whereas TCN produces a smoother and more consistent evolution of representations as the conditioning level increases.}
    \vspace{0.1in}
    \begin{tabular}{c|ccccccc}
    \toprule
    Embedding Diff. $\backslash$  Layer Index
    & 0 & 1 & 2 & 3 & 4 & 5 & 6 \\
    \midrule
    $\|\mathcal{F}_{\theta_1^*}(\mathbf{x})-\mathcal{F}_{\theta_0^*}(\mathbf{x})\|$ & 0.00 & 697.79 & 881.70 & 600.18 & 259.36 & 329.13 & 52.51 \\
    $\|\mathcal{F}_{\theta_2^*}(\mathbf{x})-\mathcal{F}_{\theta_0^*}(\mathbf{x})\|$ & 0.00 & 719.70 & 898.92 & 592.84 & 274.64 & 349.37 & 64.08 \\
    $\|\mathcal{F}_{\theta_3^*}(\mathbf{x})-\mathcal{F}_{\theta_0^*}(\mathbf{x})\|$ & 0.00 & 716.30 & 929.93 & 569.05 & 284.89 & 366.81 & 80.67 \\
    $\|\mathcal{F}_{\theta_4^*}(\mathbf{x})-\mathcal{F}_{\theta_0^*}(\mathbf{x})\|$ & 0.00 & 680.91 & 889.22 & 540.89 & 266.21 & 368.62 & 98.77 \\
    \midrule
    $\|\mathcal{T}_{\epsilon_1}\circ\mathcal{F}_{\theta^*}(\mathbf{x})-\mathcal{T}_{\epsilon_0}\circ\mathcal{F}_{\theta^*}(\mathbf{x})\|$ & 0.00 & 70.41 & 151.66 & 82.40 & 38.74 & 45.21 & 17.59 \\
    $\|\mathcal{T}_{\epsilon_2}\circ\mathcal{F}_{\theta^*}(\mathbf{x})-\mathcal{T}_{\epsilon_0}\circ\mathcal{F}_{\theta^*}(\mathbf{x})\|$ & 0.00 & 98.56 & 206.67 & 118.64 & 57.27 & 80.44 & 29.33 \\
    $\|\mathcal{T}_{\epsilon_3}\circ\mathcal{F}_{\theta^*}(\mathbf{x})-\mathcal{T}_{\epsilon_0}\circ\mathcal{F}_{\theta^*}(\mathbf{x})\|$ & 0.00 & 133.29 & 276.14 & 167.54 & 83.72 & 153.47 & 53.42 \\
    $\|\mathcal{T}_{\epsilon_4}\circ\mathcal{F}_{\theta^*}(\mathbf{x})-\mathcal{T}_{\epsilon_0}\circ\mathcal{F}_{\theta^*}(\mathbf{x})\|$ & 0.00 & 182.45 & 372.92 & 233.11 & 120.40 & 246.71 & 86.33 \\
    \bottomrule
    \end{tabular}
    \label{tab:embed_difference}
\end{table}

\subsection{Clean-level Sampling Bias}
\label{sec:clean_bias_ablation}
We use a clean-biased noise distribution
$\mathcal{P} = p \cdot \delta_{\epsilon_0} + (1-p)\,\mathrm{Unif}\{\epsilon_1,\ldots,\epsilon_N\}$,
with $p=\frac{M}{M+N}$, where $M$ controls the sampling weight on clean examples. A larger $M$ emphasizes clean training and improves clean accuracy, while a smaller $M$ places more weight on adversarial levels and favors robustness. As shown in Table~\ref{tab:ablation_clean_bias}, increasing $M$ generally improves clean accuracy on both CIFAR10 and CIFAR100, but does not consistently improve robustness at larger perturbation budgets. This suggests that a moderate clean bias achieves the best overall trade-off between clean and adversarial performance.

\begin{table}[h!]
    \centering
    \caption{Ablation study on the clean-bias coefficient in the training noise distribution. We vary the parameter $M$ in $p=\frac{M}{M+N}$ to control the probability of sampling the clean noise level during training. Larger $M$ increases the clean bias and generally improves clean accuracy, while the best overall trade-off between clean and adversarial performance is achieved at an intermediate bias level.}
    \vspace{0.1in}
    \setlength{\tabcolsep}{3pt} 
    \renewcommand{\arraystretch}{1.3}
\resizebox{1.0\textwidth}{!}{%
    \begin{tabular}{c|c|ccccc|cccccc}
    \toprule
    &\textbf{Backbone}&\multicolumn{5}{c|}{\textbf{ResNet18}}&\multicolumn{5}{c}{\textbf{WideResNet-28-10}}\\
    \midrule
\textbf{Dataset}&\textbf{$M: p=\frac{M}{M+N}$}&  \textbf{0/255} &\textbf{1/255}&\textbf{2/255}&\textbf{4/255}&\textbf{8/255}&\textbf{0/255} &\textbf{1/255}&\textbf{2/255}&\textbf{4/255}&\textbf{8/255}
\\\midrule
&1&92.08 & 86.44 & 80.44 & 66.69 & 43.45&94.19 & 88.98 & 82.62 & 69.40 & 48.27 \\
\multirow{2}{*}{\textbf{CIFAR10}} &2&92.77 & 87.01 & 80.58 & 66.83 & 44.85&94.56 & 88.49 & 82.32 & 68.87 & 46.63\\
&3&93.85 & 87.46 & 80.49 & 66.36 & 45.16&94.35 & 88.21 & 81.06 & 66.98 & 45.36\\
&4&93.62 & 87.11 & 80.3 & 66.23 & 44.23&94.91 & 87.99 & 80.76 & 66.15 & 43.71\\
    \midrule
&1&64.49 & 55.59 & 47.4 & 34.48 & 17.95&73.31 & 62.73 & 53.81 & 38.79 & 21.81\\
\multirow{2}{*}{\textbf{CIFAR100}} &2&66.22 & 56.44 & 47.95 & 34.19 & 17.57&74.86 & 62.57 & 52.70 & 38.42 & 21.50\\
&3&67.97 & 57.28 & 47.92 & 34.19 & 18.18&75.42 & 62.73 & 52.12 & 37.76 & 22.21\\
&4&68.68 & 57.08 & 47.59 & 34.25 & 17.99&76.52 & 61.85 & 51.28 & 37.71 & 21.14\\
    \bottomrule
    \end{tabular}
}
    \label{tab:ablation_clean_bias}
\end{table}



\end{document}